\pdfoutput=1
\PassOptionsToPackage{table}{xcolor}
\documentclass[11pt]{article}

\usepackage[preprint]{acl}

\usepackage{times}
\usepackage{latexsym}
\usepackage{amsmath}
\usepackage{booktabs}
\usepackage{graphicx}
\usepackage{url}
\usepackage{multirow}
\usepackage[table]{xcolor}
\usepackage{subcaption}
\usepackage{tikz}
\usepackage{pgfplots}
\usepackage{pgf-pie}
\usepackage{pgfplotstable}
\usepackage{float}
\usepackage{diagbox}
\usepackage{array}
\usepackage{stfloats}

\usepackage[T1]{fontenc}

\usepackage[utf8]{inputenc}

\usepackage{microtype}

\usepackage{inconsolata}

\usepackage{graphicx}

\title{Bel Esprit: Multi-Agent Framework for Building AI Model Pipelines}

\author{Yunsu Kim \quad AhmedElmogtaba Abdelaziz \quad Thiago Castro Ferreira \\
  {\bf Mohamed Al-Badrashiny} \quad {\bf Hassan Sawaf} \\
  aiXplain, Inc.\\
  Los Gatos, CA, USA\\
  \texttt{\{firstname.lastname\}@aixplain.com}}

\begin{document}
\maketitle
\begin{abstract}
As the demand for artificial intelligence (AI) grows to address complex real-world tasks, single models are often insufficient, requiring the integration of multiple models into pipelines.
This paper introduces \emph{Bel Esprit}, a conversational agent designed to construct AI model pipelines based on user requirements.
Bel Esprit uses a multi-agent framework where subagents collaborate to clarify requirements, build, validate, and populate pipelines with appropriate models.
We demonstrate its effectiveness in generating pipelines from ambiguous user queries, using both human-curated and synthetic data.
A detailed error analysis highlights ongoing challenges in pipeline building.
Bel Esprit is available for a free trial at \url{https://belesprit.aixplain.com}\footnote{Demo video: \url{https://youtu.be/3KFSvrOPObY}}.
\end{abstract}

\section{Introduction}
\label{sec:intro}

\begin{figure}[!t]
\footnotesize\raggedright\noindent\textbf{Query}: I want to dub my video clip in French, German, and Spanish
  \includegraphics[width=\columnwidth]{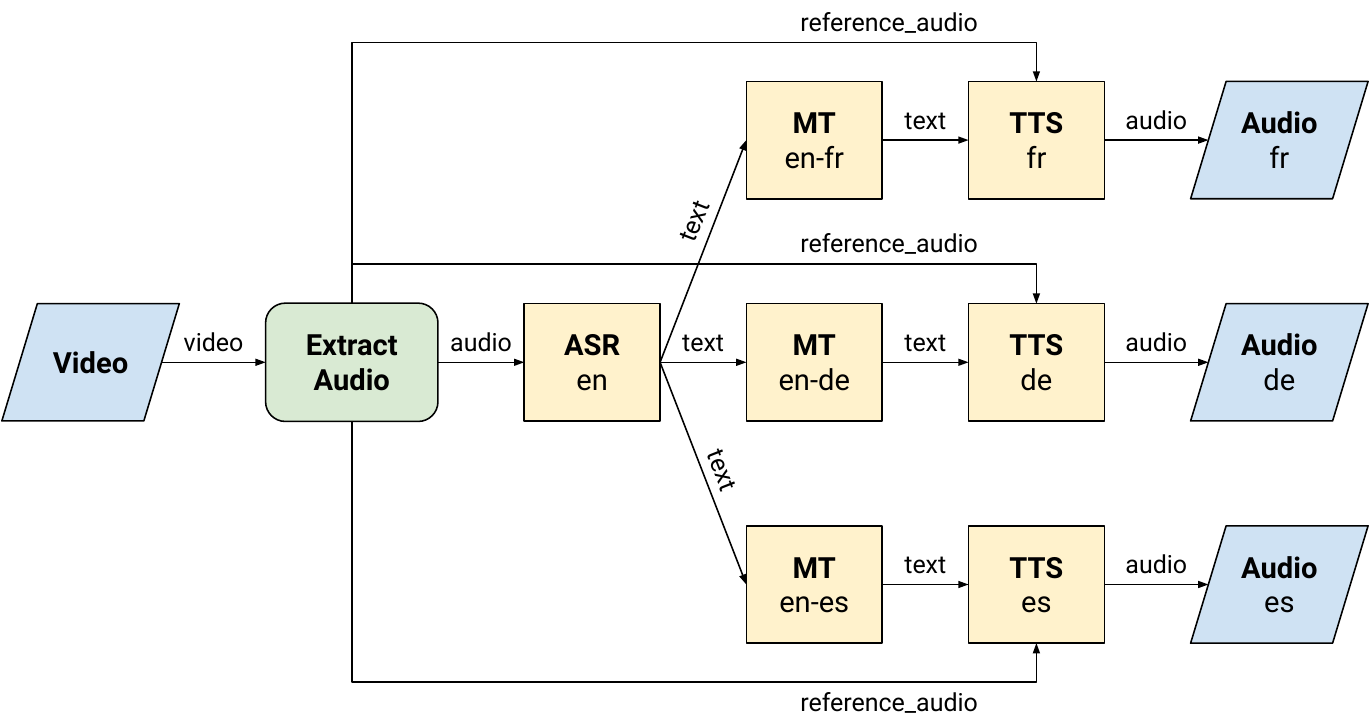}
  \caption{Query and model pipeline for multilingual video dubbing.}
  \vspace*{-0.5em}
  \label{fig:intro:example}
\end{figure}

A single AI model is often insufficient for complex tasks, especially with multiple inputs or outputs, e.g., multimodal content moderation or multilingual video dubbing (Figure \ref{fig:intro:example}).
Such tasks can be better addressed by integrating different models; by constructing a pipeline of interconnected models, we can automate intermediate steps and facilitate seamless task transitions.
This approach, known as cascading models into a pipeline, has been widely used in applications like speech translation \cite{ney1999speech,matusov2009combining} and voice conversion \cite{wu2018nu,huang2020sequence}.

This paper presents \emph{Bel Esprit}\footnote{French for ``beautiful mind''}, a conversational assistant that implements sophisticated pipeline solutions composed of diverse AI models.
Here are our main contributions:
\begin{itemize}\itemsep0em
    \item We formally define the task of model pipeline building as a graph generation problem involving scientific reasoning.
    \item We design a multi-agent framework that systematically enhances pipeline quality and alignment with user intent.
    \item We establish a rigorous evaluation scheme for pipeline building, including a data preparation protocol and automatic metrics.
\end{itemize}

\section{Related Work}
\label{sec:related}

\noindent\textbf{Automated Machine Learning}\hspace{1em} Efforts to simplify machine learning for non-experts have focused on automating model selection \cite{kotthoff2017auto}, neural architecture search \cite{jin2019auto,zimmer2021auto}, hyperparameter tuning \cite{bischl2023hyperparameter}, and ensembling \cite{erickson2020autogluon,shchur2023autogluon}: mainly aiming to train a single model for atomic tasks.
In contrast, Bel Esprit does not train models but assembles off-the-shelf models into pipelines, integrating various AI components for more complex tasks.
\vspace{0.5em}

\noindent\textbf{Agentic Workflow Generation}\hspace{1em}
Modern AI agents use multiple tools and subagents to break down complex tasks into subtasks and assign tools accordingly \cite{xi2023rise,wang2024survey}.
Existing workflow generation methods largely focus on writing LLM prompts for a few general agents or ordering simple utility functions, with evaluations limited to classical reasoning tasks like math, coding, or QA \cite{chen2023autoagents,zeng2023flowmind,li2024autoflow,zhuge2024gptswarm,zhang2024aflow,hu2024automated,niu2025flow}.

Bel Esprit expands this scope by integrating >70 AI functions across modalities (Appendix \ref{sec:app:function}) and devising tools for missing functionalities.
It ensures pipeline reliability through conversational requirement clarification and formal graph-based verification.
Also, the generated pipelines can serve as advanced tools within agents, reducing redundant planning and accelerating recurring tasks \cite{qian2023creator,wang2024voyager,cai2024large}.





\section{Task Definition}
\label{sec:task}

\begin{figure}[!t]
  \includegraphics[width=\columnwidth]{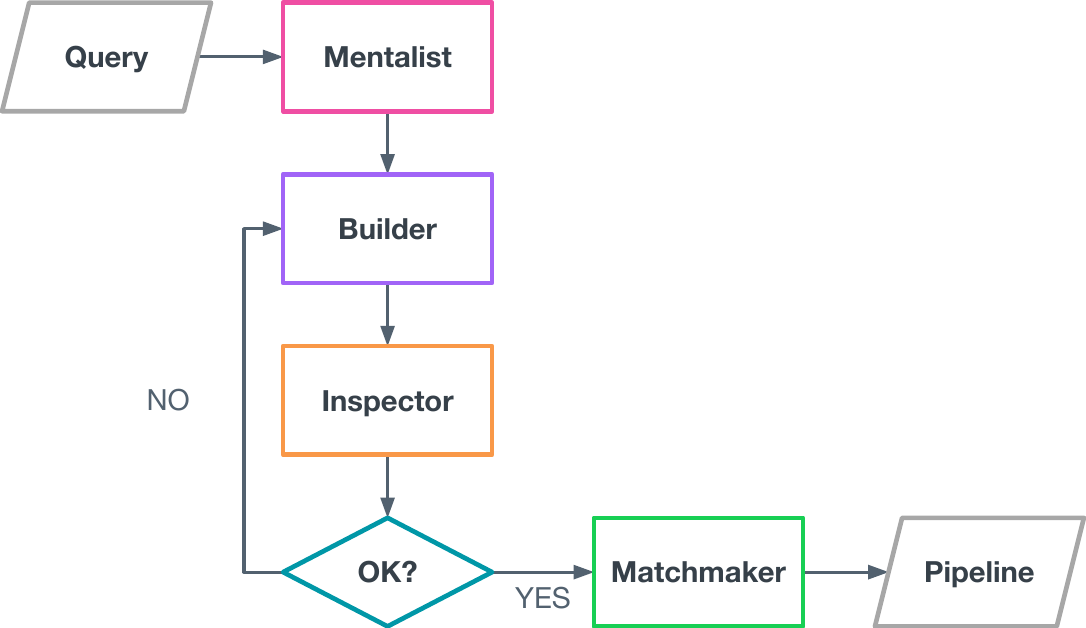}
  \caption{Agentic flow of Bel Esprit.}
  \label{fig:frame:flow}
\end{figure}

Pipeline generation is a structured prediction task, where the input is a user query describing a computational task, and the output is a pipeline of AI functions to solve it.
Each AI function may have parameters, e.g., language in speech recognition.
The final output is basically a graph, with nodes representing inputs/outputs/functions, and edges denoting the data flow between them.
To enhance the functionality of a pipeline, we introduce three special node types:

\begin{itemize}\itemsep0em
    \item \textbf{Router}: Directs the input data to different paths based on its modality.
    \item \textbf{Decision}: Sends data to different paths according to specific input values.
    \item \textbf{Script}: Executes an arbitrary task by running Python code.
\end{itemize}

Pipeline generation can be viewed as deductive reasoning where the AI functions exist as \emph{premises} about data \emph{entities} \citep{yu2023natural}.
Each premise conveys scientific knowledge from specific input to output.
Given a user query as a new comprehensive \emph{conclusion}, the objective is to find a reasoning path comprising multiple premises \citep{saha2020prover,creswell2022selection,saparov2022language}.

\section{Framework}
\label{sec:frame}

In this work, we use an LLM to process user queries and generate pipeline structures through guided prompts.
Instead of producing the pipeline in a single step, the framework follows a flow of multiple subagents (Figure \ref{fig:frame:flow}).
The process begins with \emph{Mentalist}, followed by \emph{Builder}, which creates an initial pipeline.
This pipeline is then reviewed by \emph{Inspector}.
If the review fails, it loops back to Builder for revisions until an error-free pipeline is generated or the maximum iteration limit is reached.
Once the pipeline passes inspection, it proceeds to \emph{Matchmaker}, completing the final pipeline.

\subsection{Mentalist}
\label{sec:frame:mentalist}

Mentalist is the agent responsible for interacting with the user and analyzing their requirements.
\begin{figure}[!t]
  \includegraphics[width=\columnwidth]{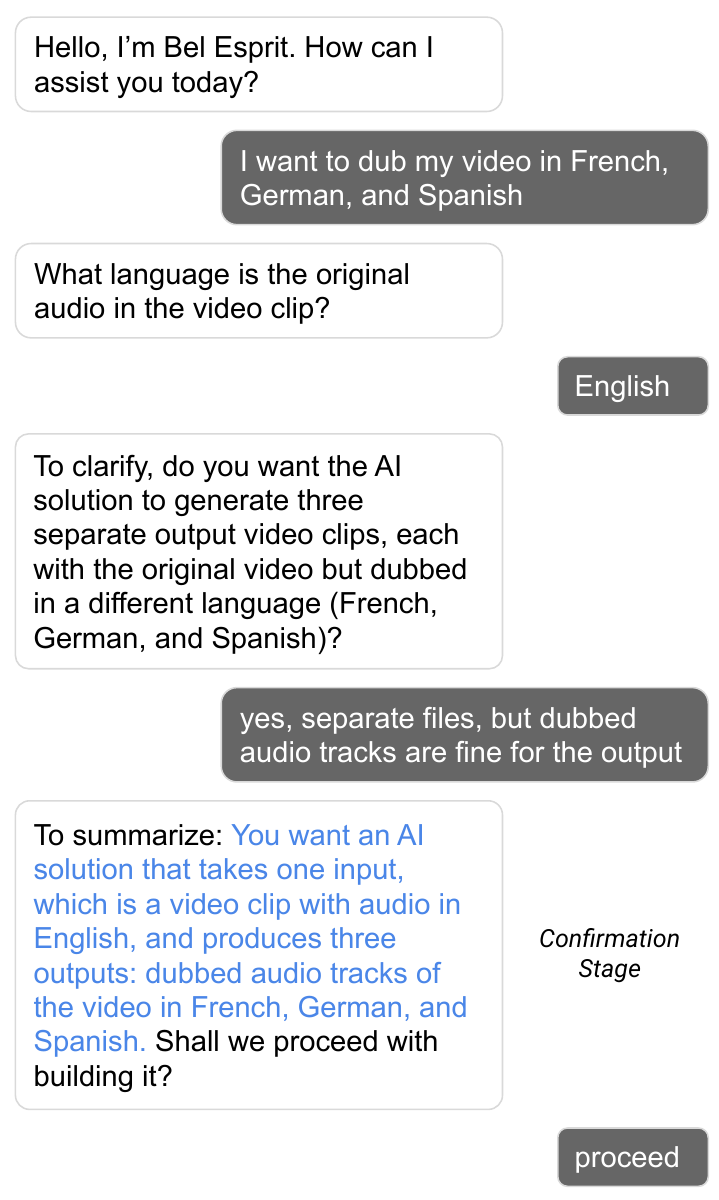}
  \caption{Example conversation between Mentalist and a user. The refined query is colored in blue.}
  \label{fig:frame:mentalist:convo}
\end{figure}

\subsubsection{Query Clarifier}
\label{sec:frame:mentalist:query}

User queries are often too ambiguous to build a correct solution.
For example, they may lack detailed context, such as how ``risk'' is defined in a risk management system, or omit data properties, like the language of the input text.
\emph{Query Clarifier}, a chat interface, converts potentially ambiguous user queries into fully developed solution specifications.
It identifies missing information and prompts the user to fill in the gaps.
Once all necessary details are gathered, the system summarizes the conversation into a refined query that clearly outlines the solution’s inputs and outputs, along with their modalities and relationships (Figure \ref{fig:frame:mentalist:convo}).

\subsubsection{Specification Extractor}
\label{sec:frame:mentalist:spec}
\begin{table}[!t]
  \centering
  \scalebox{0.9}{
  \begin{tabular}{cccc}
  \toprule
  & \textbf{Name} & \textbf{Modality} & \textbf{Language} \\ 
  \midrule
  Input  & Video file  & Video  & English \\
  \midrule
  \multirow{3}{*}{Output} & Audio track 1 & Audio & French \\
   & Audio track 2 & Audio & German \\
   & Audio track 3 & Audio & Spanish \\
  \bottomrule
  \end{tabular}}
  \caption{Specification example.}
  \label{tab:frame:mentalist:spec}
\end{table}

After the user confirms the clarified query, \emph{Specification Extractor} extracts its technical details like name, modality, and required parameters for each input and output (Table \ref{tab:frame:mentalist:spec}).
Such structured information offers clear guidance on which input and output nodes must be included, providing a strong foundation for constructing the intermediate flows; relying solely on long natural language queries often results in errors when building a solution.

\subsubsection{Attachment Matcher}
\label{sec:app:attach}

We found that many users begin by attaching a file, e.g., ``I want to work with this text file to extract named entities and identify grammatical errors.''
Once a solution is generated, users need to know which input node in the pipeline graph corresponds to the attached file.
While matching is straightforward for only a single input node, it becomes challenging when there are multiple input nodes, especially when some share the same modality.

In such cases, semantic analysis of the conversation is necessary to determine the specific characteristics of each input.
Files may also be attached mid-conversation, with contextual clues before and after the attachment providing critical information for accurate matching.
\emph{Attachment Matcher} detects these associations and assigns each attached file to the appropriate input node.
Note that file names themselves are not passed to the builder, as they may not be directly relevant to the solution.

\begin{figure}[!t]
  \includegraphics[width=\columnwidth]{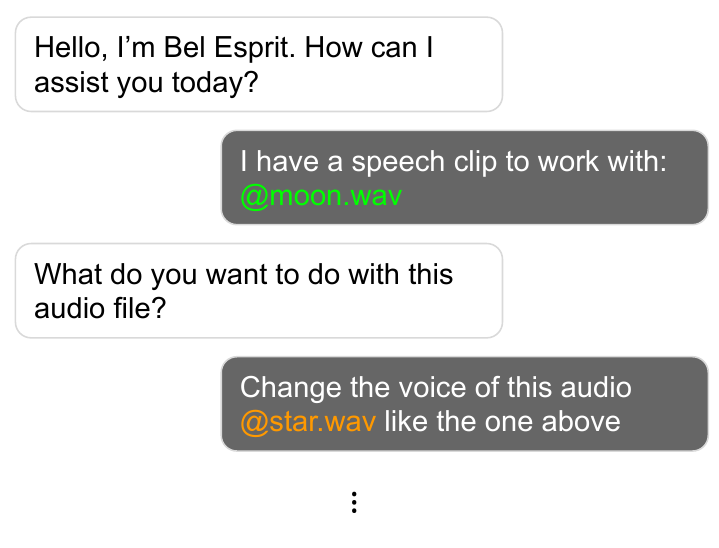}
  \centering
  \scalebox{0.9}{
  \begin{tabular}{ccc}
  \toprule
  & \textbf{Name} & \textbf{Attachment} \\
  \midrule
  \multirow{2}{*}{Input} & Input speech & \textsf{\small\color{orange} @star.wav} \\
  & Reference voice & \textsf{\small\color{green} @moon.wav} \\
  \midrule
  Output & Converted speech & N/A \\
  \bottomrule
  \end{tabular}}
  \caption{Attachment matching example.}
  \label{fig:frame:mentalist:attach}
\end{figure}

\subsection{Builder}
\label{sec:frame:builder}

Builder constructs the pipeline graph based on the refined query (Section \ref{sec:frame:mentalist:query}) and the extracted specification (Section \ref{sec:frame:mentalist:spec}).
Builder is an LLM prompted with information on data types, function identifiers, node types, and graph constraints (Appendix \ref{sec:app:graph}).
Given the complexity of this task, a few example pipelines are included in the prompt to guide the generation process \citep{brown2020language}.
Builder’s output can be in any structured format, such as DOT or JSON.

\subsubsection{Chain-of-Branches}
\label{sec:frame:builder:chain}

Building a large graph in a single step is highly challenging.
Generating token sequences in structured formats often leads to issues like hallucination and loss of consistency within the structure \cite{poesia2022synchromesh,beurer2024guiding,tam2024let}.
Inspired by the chain-of-thought \cite{wei2022chain}, we decompose the solution graph into distinct branches.
Each branch represents a path from one or more input nodes to an output node; a pipeline with $N$ output nodes will have $N$ branches.
These branches can be standalone solutions to subproblems derived from the user query.
New branches can often reuse nodes from existing branches, reducing the number of totally new nodes to be generated for each branch.

\begin{figure}[!t]
  \includegraphics[width=\columnwidth]{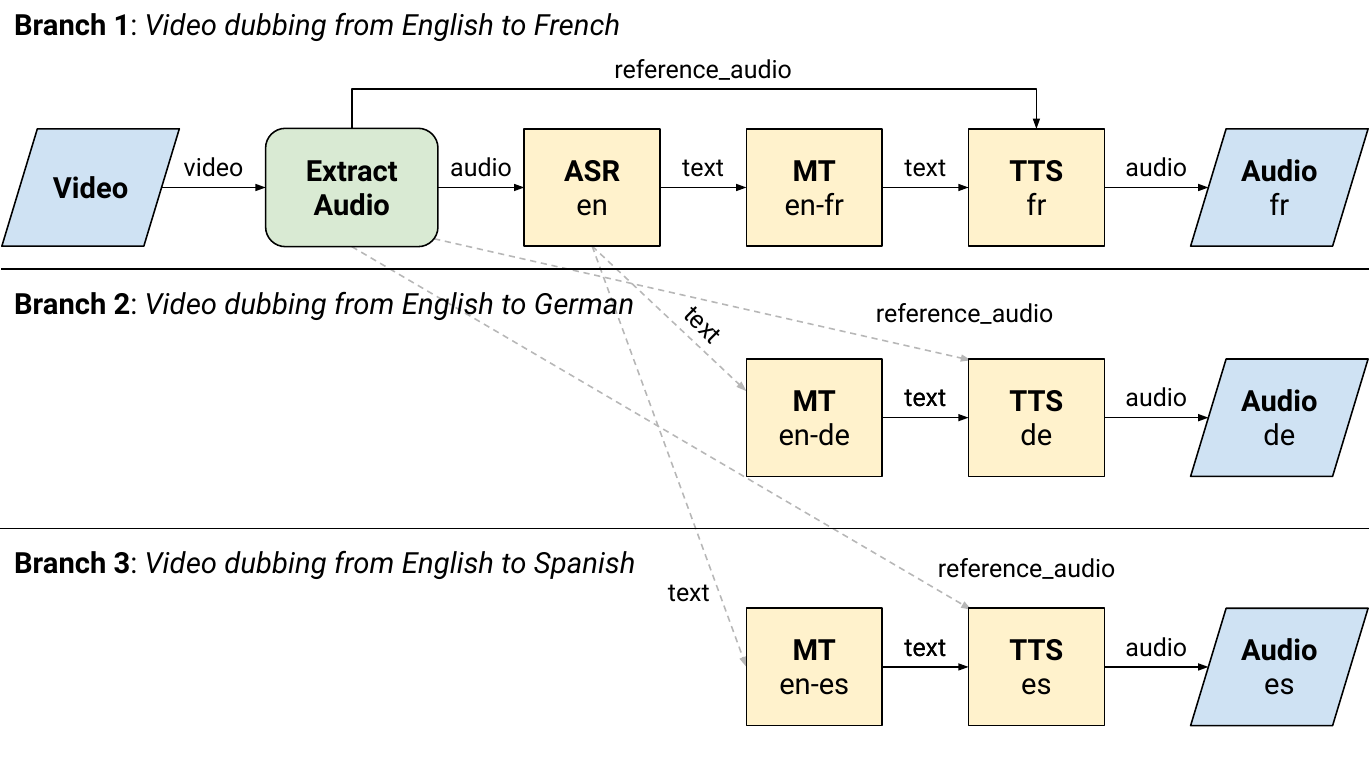}
  \caption{Example of generation using chain-of-branches. Gray dashed arrows indicate connections to previously generated nodes in existing branches. The final pipeline is in Figure \ref{fig:intro:example}.}
  \label{fig:frame:builder:chain}
\end{figure}

We prompt the LLM to generate one branch at a time, completing all nodes and edges for that branch before moving to the next (Figure \ref{fig:frame:builder:chain}).
At each branch, we instruct the model to generate a brief comment to clarify the subproblem it addresses, ensuring the boundaries between branches.

\subsection{Inspector}
\label{sec:frame:inspector}

LLMs are particularly vulnerable to errors in scientific reasoning on lengthy contexts \cite{ahn2024large,ma2024sciagent}.
Even with a clarified query, errors may still occur due to the solution complexity.
Similarly to critic models for LLM outputs \cite{ke2023critiquellm,xu2024llmrefine,gou2024critic}, we developed \emph{Inspector}, which analyzes the builder's output to identify errors in both the graph structure and semantic alignment with user requirements.

\subsubsection{Syntax}
\label{sec:frame:inspector:syntax}

First, we assess the structural integrity of the generated graph, independent of its intended function.
We check violations of graph constraints (Appendix \ref{sec:app:graph}), often due to improper node connections.

\begin{figure}[!t]
  \centering
  \begin{subfigure}[b]{\columnwidth}
    \includegraphics[width=\columnwidth]{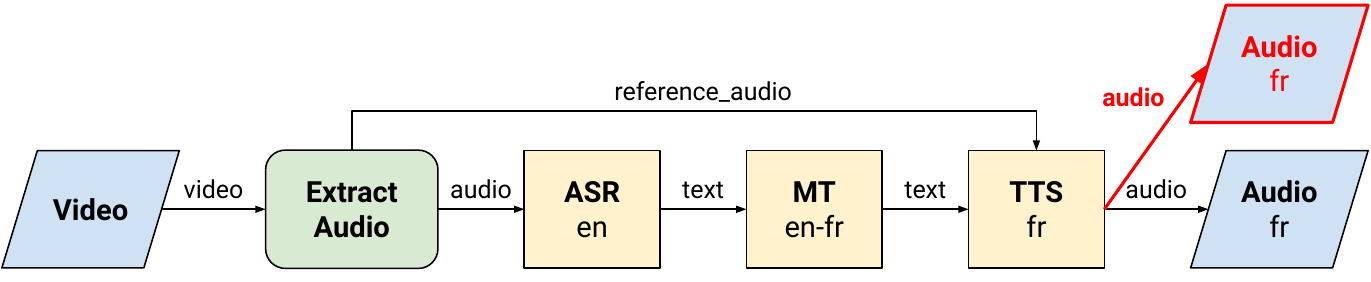}
    \caption{Mechanically correctable}\vspace{0.5em}
    \label{fig:frame:inspector:syntax:self-fix}
  \end{subfigure}
  \begin{subfigure}[b]{\columnwidth}
    \includegraphics[width=\columnwidth]{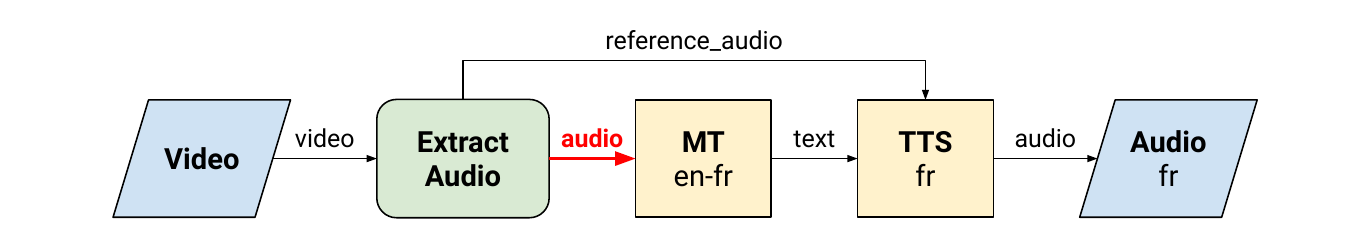}
    \caption{LLM-assisted correctable}
    \label{fig:frame:inspector:syntax:llm-fix}
  \end{subfigure}
  \caption{Example of syntax errors (highlighted in red).}
  \label{fig:frame:inspector:syntax}
\end{figure}

Some violations can be mechanically corrected immediately upon detection.
Figure \ref{fig:frame:inspector:syntax:self-fix} illustrates such a case in generating Branch 1 of Figure \ref{fig:frame:builder:chain}.
The output from a function node should connect to one output node, but multiple output nodes are linked to the same function output.
This often arises when the user specifies multiple outputs in the solution.
Such errors can be resolved by retaining only one output node and removing the duplicates.

Figure \ref{fig:frame:inspector:syntax:llm-fix} illustrates an example where no simple correction is feasible.
The machine translation (MT) node requires text input, yet audio extracted from a video input is routed directly to it.
Resolving this modality mismatch involves either locating an existing node producing the necessary text output or creating a new node for the required transformation.
Such complex corrections require an LLM to reconstruct the graph (Section \ref{sec:frame:builder}).

\begin{figure}[!t]
  \includegraphics[width=\columnwidth]{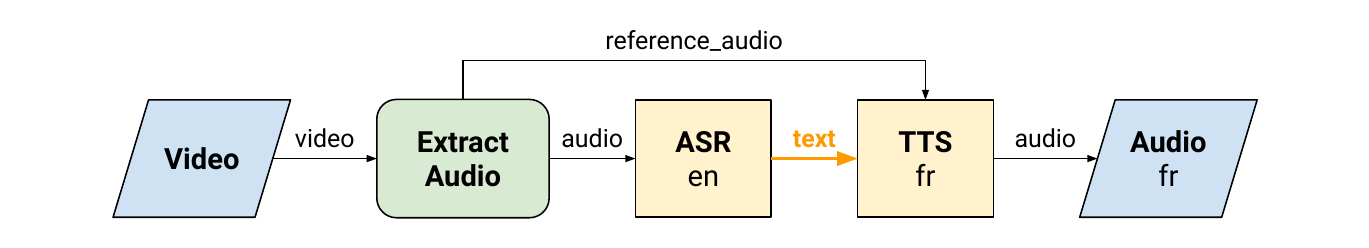}
  \caption{Example of semantic errors in a branch (highlighted in orange).}
  \label{fig:frame:inspector:semantics}
\end{figure}

\subsubsection{Semantics}
\label{sec:frame:inspector:semantics}

Next, we verify whether the graph semantically fulfills the user requirements.
For each branch, we provide an LLM with a natural language summary that lists the nodes sequentially, outlining the path and its context within the pipeline.
The LLM then identifies the corresponding requirements in the specification (Section \ref{sec:frame:mentalist:spec}) and flags any unmatched or missing steps in the branch path.

Figure \ref{fig:frame:inspector:semantics} shows an example where the branch passes structural checks but fails in semantic alignment.
In this case, the English transcription is routed directly to a French text-to-speech (TTS) node, assuming the same text modality suffices for synthesis; the builder overlooked the necessary translation step, resulting in a mismatch between the automatic speech recognition (ASR) output language and the intended TTS language.


\subsection{Matchmaker}
\label{sec:frame:matchmaker}

A pipeline from the Builder specifies only the data flow without assigning specific models to function nodes.
\emph{Matchmaker} gathers any additional information about the model selection in the query and finds the model that best align with the user's preferences, e.g., the latest MT model from Google or an ASR model specialized in medical domain.
When no specific preference is provided, Matchmaker defaults to a predefined model choice.

If a node requires a task for which no suitable model exists---often due to a complex user query or gaps in the platform’s model library---Matchmaker employs the following fallback strategies.


\subsubsection{Generic Nodes}
\label{sec:frame:matchmaker:node}

\begin{figure}[!t]
\footnotesize\raggedright\noindent\textbf{Query}: I want to understand English news clips more easily\\[1em]
  \includegraphics[width=\columnwidth]{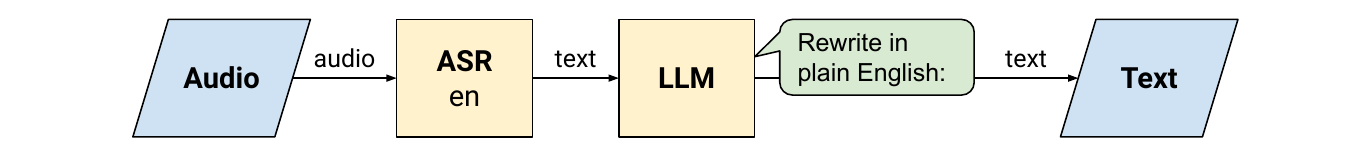}
  \caption{Example pipeline using a generic node.}
  \label{fig:frame:matchmaker:node}\vspace{0.5em}
\end{figure}

Recent LLMs can perform generic tasks beyond their specific training when given a clear prompt \cite{brown2020language,wei2022emergent}.
For unavailable AI functions, we insert a custom LLM node with a prompt derived from the relevant part of the user query (Figure \ref{fig:frame:matchmaker:node}).
This approach is useful for tasks like domain mixing or creative writing, where specialized models are scarce.

\begin{figure}[!t]
\footnotesize\raggedright\noindent\textbf{Query}: If I give you a summary, extend it to a long article; if it's an article, then summarize it.\\[1.2em]
  \includegraphics[width=\columnwidth]{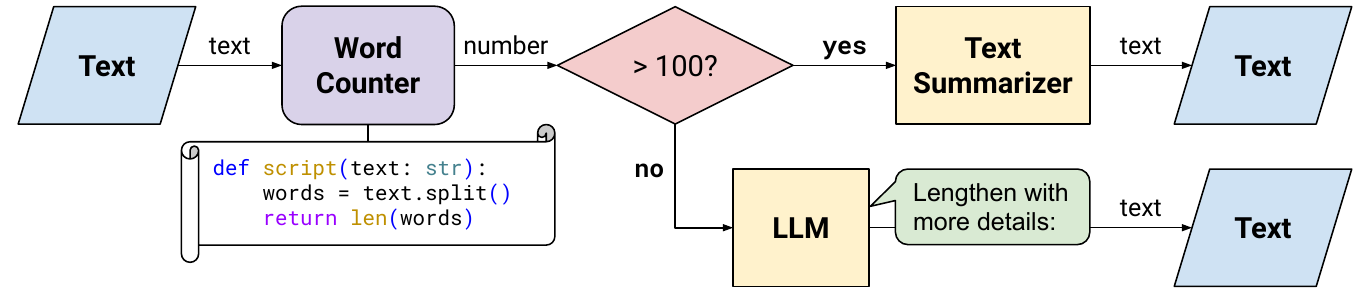}
  \caption{Example pipeline with a script node.}
  \label{fig:frame:matchmaker:script}
\end{figure}

\subsubsection{Script Generator}
\label{sec:frame:matchmaker:script}

Some nodes are designated not for AI tasks but for simpler functions, such as counting words or extracting text from a PDF: a short script is sufficient (Figure \ref{fig:frame:matchmaker:script}).
In such cases, we use an LLM to generate scripts; we begin by providing a script template that defines the input/output and their modalities, allowing the LLM to complete the method part based on the task description.

\section{Experiments}
\label{sec:exp}

To evaluate pipeline generation, we prepared query-pipeline pairs with evaluation metrics.

\subsection{Data}
\label{sec:exp:data}

\noindent\textbf{Manual creation}\hspace{1em}
Given the high-level scientific nature of the task, we recruited five AI solution engineers at aiXplain, Inc. to create 82 realistic tasks and their corresponding pipelines.
Each pipeline was then reviewed and, if necessary, revised by at least one other expert.
\vspace{0.5em}

\noindent\textbf{Structured synthesis with human correction}\hspace{1em}
To scale data collection, we automated the initial pipeline creation using rule-based expansion: nodes in a pipeline are expanded by adding others that match the input-output specifications.
Starting with one or more input nodes, we constructed a tree-like structure that can branch into multiple output nodes.
To manage complexity, we parameterized the number of AI function nodes and restrict each node to have a maximum of two children.

An LLM generates specifications and clear queries that enumerate the inputs and outputs.
To simulate realistic user interactions, we then synthesized an initial user query by intentionally introducing ambiguity into the LLM prompt.
In this way, we synthesized 500 data entries, retaining 359 after human review.
\vspace{0.5em}

\noindent In total, we curated a dataset of 441 pipelines.
For further details of the data, see Appendix \ref{sec:app:data}.

\subsection{Metrics}
\label{sec:exp:metrics}

\noindent\textbf{Exact Match (EM)}\hspace{1em}First, we count cases where the generated pipeline exactly matches the reference pipeline.
Two nodes are considered a match if their types are identical and, if applicable, their functions and parameters are the same.
For LLM nodes, we match prompts based on cosine similarity of their sentence embeddings, with a threshold of $0.5$.
For script nodes, we consider two code snippets a match if an LLM determines they perform the same task.
Edges are matched if they connect the same source and target nodes with identical parameters.
Determining such an exact match (EM) requires solving the graph isomorphism problem.
To implement this, we adapted the VF2 algorithm \cite{cordella2004sub} to account for our problem.
\vspace{0.5em}

\noindent\textbf{Graph Edit Distance (GED)}\hspace{1em} In our initial study, we found that many non-matching pipelines differ only slightly, typically by a single node or edge.
Assigning a full penalty to such cases is too severe, as EM fails to capture incremental improvements.
Therefore, we adopt graph edit distance (GED), which counts the number of edit operations---insertion, deletion, or substitution of nodes or edges---required to convert the generated graph to its reference.
We apply the same matching conditions for nodes and edges as used in EM.

We used the depth-first GED algorithm \cite{abu2015exact} implemented in NetworkX \cite{hagberg2008exploring}.
The edit operations have an equal weight of $1.0$ for simplicity.
We limited the running time for each pipeline pair to 60 seconds on Macbook Pro 2023 (with M2 Pro).

\subsection{Models}
Mentalist's query clarifier (Section \ref{sec:frame:mentalist:query}) and Builder (Section \ref{sec:frame:builder}) utilize GPT-4o \cite{openai2024gpt}, while the rest of the framework, including data synthesis and evaluation, relies on the Llama 3.1 70B model \cite{dubey2024llama} when LLM assistance is required.
Prompt similarity is computed using the all-MiniLM-L6-v2 model of Sentence Transformers \cite{reimers2019sentence}.

\begin{table*}[!ht]
\centering
\scalebox{0.9}{
\begin{tabular}{lcccccc}
\toprule
 & \multicolumn{2}{c}{GPT-4o} & \multicolumn{2}{c}{Llama 3.1 405B} & \multicolumn{2}{c}{Llama3.1 70B} \\ 
\cmidrule(lr){2-3} \cmidrule(lr){4-5} \cmidrule(lr){6-7}
Framework setup & EM [\%] & GED [\%] & EM [\%] & GED [\%] & EM [\%] & GED [\%] \\
\midrule
Builder & 15.7 & 65.1 & 13.6 & 71.7 & 14.1 & 70.7 \\ 
\hspace{0.01em} + Query clarifier & 25.1 & 44.4 & 21.5 & 52.8 & 19.0 & 54.4 \\ 
\hspace{0.5em} + Specification extractor & 26.0 & 41.4 & 21.9 & 52.6 & 21.1 & 52.7 \\ 
\hspace{1.0em} + Chain-of-branches & 25.2 & 40.3 & 21.9 & 52.6 & 19.0 & 53.9 \\ 
\hspace{1.5em} + Syntactic inspector & 25.6 & 38.3 & 22.7 & 48.2 & 19.4 & 49.8 \\ 
\hspace{2.0em} + Semantic inspector & 25.2 & 37.0 & 20.3 & 48.9 & 19.4 & 53.9 \\ 
\bottomrule
\end{tabular}}
\caption{Pipeline generation performance across framework configurations and Builder LLMs.}
\label{tab:results:main}
\end{table*}

\subsection{Results}
\label{sec:results}

Table \ref{tab:results:main} shows the pipeline generation performance across various framework configurations.
Starting with a baseline pipeline builder, we incrementally incorporate components from Mentalist, Builder, and Inspector, achieving +9.5\% EM and -28.1\% GED overall.
For the Builder, GPT-4o outperforms open-source alternatives, with performance declining as model size decreases.
Smaller models like Llama 3.1 8B yielded unacceptable performances, with EM rates below 3\%.

Each component's contribution is evident in GED improvements for GPT-4o but less consistent for weaker models, while EM fails to capture the nuanced improvements.
As a side note, semantic inspection occasionally confuses weaker Builders, leading to unnecessary graph repetitions and sporadic performance drops.

\begin{figure}[!th]
\footnotesize\raggedright\noindent\textbf{Query}: I want to translate my speech into French and German\\[1em]
  \begin{subfigure}[b]{\columnwidth}
    \includegraphics[width=\columnwidth]{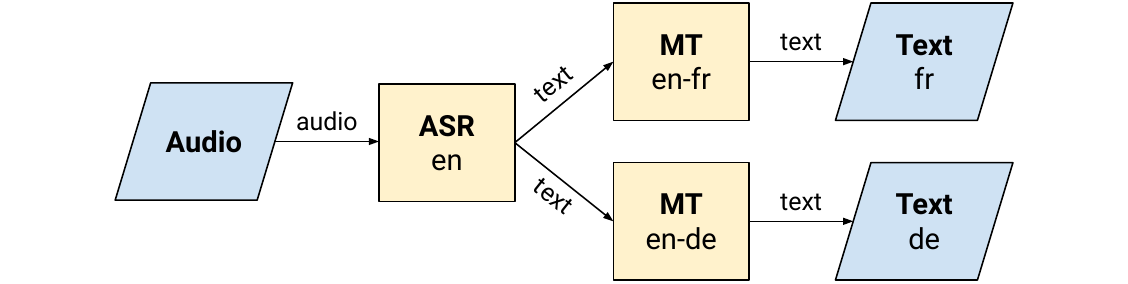}
    \caption{Builder (plain)}\vspace{0.5em}
    \label{fig:results:example:asr-en}
  \end{subfigure}
\footnotesize\raggedright\noindent\textbf{Refined Query}: The requested solution takes speech in an unknown language as input and converts it to French text. The input language will be detected automatically.\\[1em]
  \begin{subfigure}[b]{\columnwidth}
    \includegraphics[width=\columnwidth]{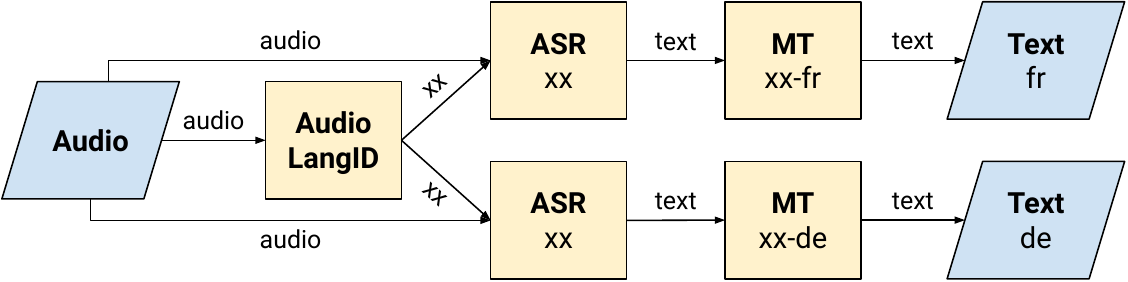}
    \caption{Mentalist + Builder (plain)}\vspace{1em}
    \label{fig:results:example:asr-repeated}
  \end{subfigure}
  \begin{subfigure}[b]{\columnwidth}
    \includegraphics[width=\columnwidth]{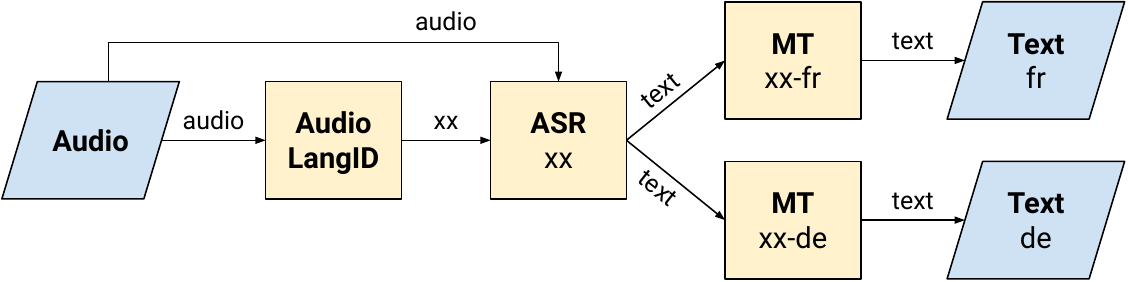}
    \caption{Mentalist + Builder (chain-of-branches)}\vspace{1em}
    \label{fig:results:example:no-srclang}
  \end{subfigure}
  \begin{subfigure}[b]{\columnwidth}
    \includegraphics[width=\columnwidth]{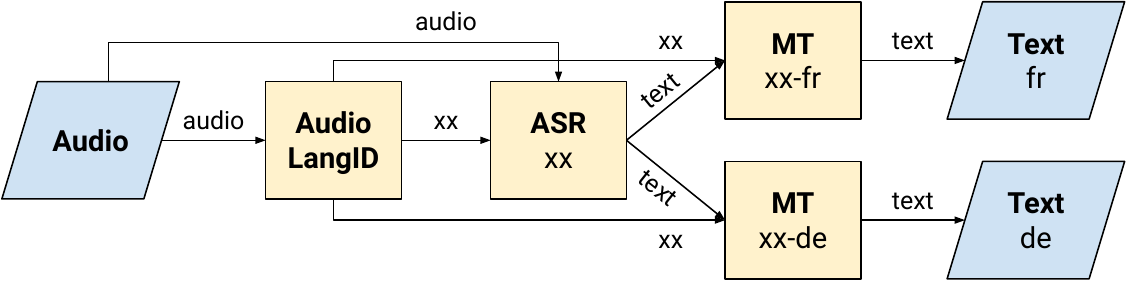}
    \caption{Mentalist + Builder (chain-of-branches) + Inspector}
    \label{fig:results:example:complete}
  \end{subfigure}
  \caption{Examples of generated pipelines across different framework configurations.}
  \vspace{-1em}
  \label{fig:results:example}
\end{figure}

\subsection{Qualitative Example}
\label{sec:exp:example}
Figure \ref{fig:results:example} illustrates an example of incremental improvements in pipeline generation.
The initial user query is ambiguous, as it does not specify the input language.
The plain Builder assumes English as the input language and generates a pipeline accordingly.
Mentalist refines the query to explicitly indicate that the input language is unknown, resulting in a pipeline that first performs language identification and passes the detected language to the ASR function.

However, this version redundantly includes separate ASR nodes for French and German outputs.
The chain-of-branches technique resolves this redundancy by generating one path at a time, enabling the reuse of the ASR node.
Despite this improvement, the MT nodes lack source language parameters.
The final configuration, which incorporates Inspector, identifies this issue and adds edges from the language identifier to the MT nodes, producing a complete and correct pipeline.

\section{Analysis}
\label{sec:analysis}

\begin{figure}[!t]
  \includegraphics[width=\columnwidth]{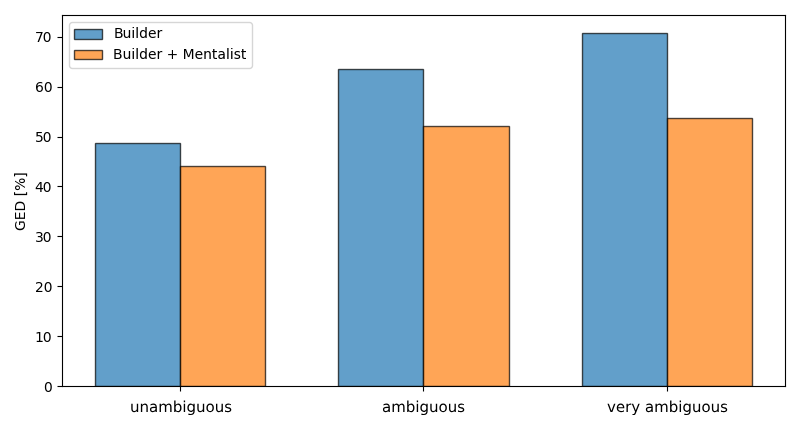}
  \vspace*{-1.3em}
  \caption{GED over increasing query ambiguity.}
  \label{fig:results:ambiguity}
\end{figure}

\noindent\textbf{Ambiguity of query}\hspace{1em}
As shown above, ambiguity in user queries is a primary factor for poor pipeline generation performance.
We used GPT-4o to rate the ambiguity of queries in three levels: unambiguous, ambiguous, and very ambiguous.
Figure \ref{fig:results:ambiguity} shows performance computed for each level; pipeline generation becomes increasingly challenging with higher ambiguity.
The Mentalist subagent significantly improves performance in such cases by clarifying missing information in queries and concretizing input and output requirements.
\vspace{0.5em}

\begin{figure}[!t]
  \includegraphics[width=\columnwidth]{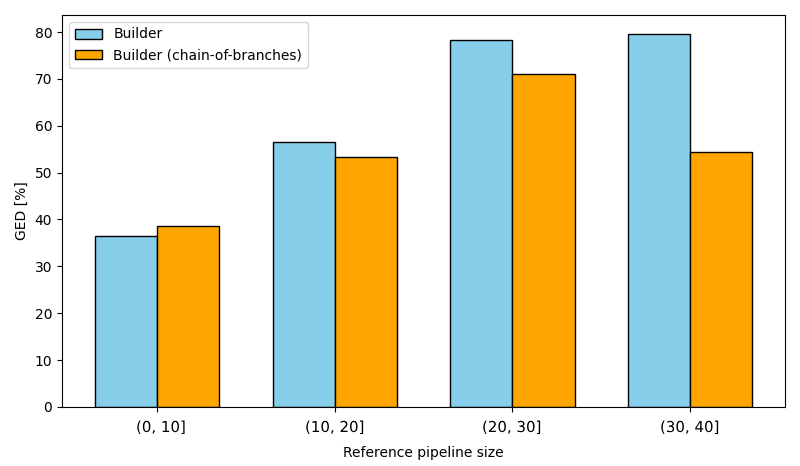}
  \vspace*{-1.5em}
  \caption{GED over increasing pipeline size.}
  \label{fig:results:size}
  \vspace{-0.5em}
\end{figure}

\noindent\textbf{Pipeline size}\hspace{1em}
We also measured performance as a function of reference pipeline size, shown in Figure \ref{fig:results:size}.
As expected, larger pipelines—--such as simultaneous processing of the same input across multiple paths---are more challenging to construct.
However, the chain-of-branches technique proves to be effective in handling these cases by breaking the graph into manageable subgraphs.
\vspace{0.5em}

\begin{figure}[!t]
  \vspace*{-1.5em}
  \includegraphics[width=\columnwidth]{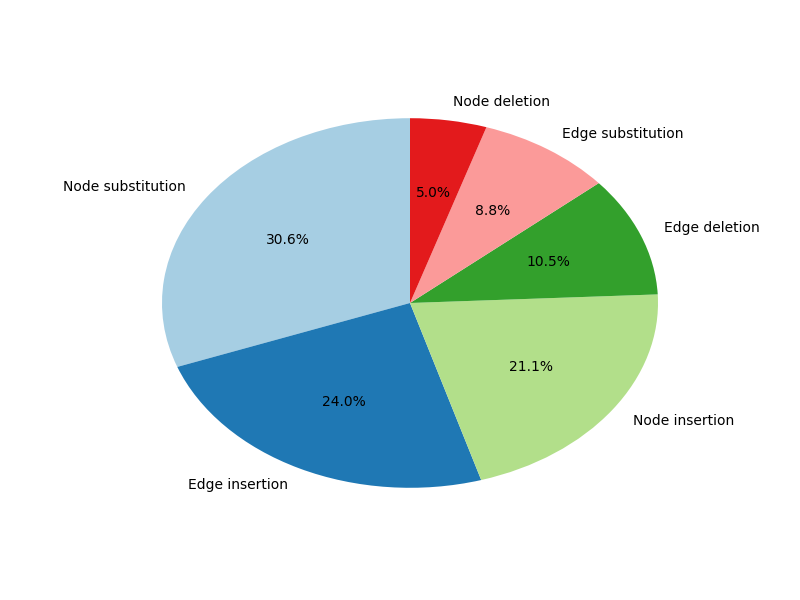}
  \vspace*{-3.5em}
  \caption{Distribution of edits required to align generated pipelines with reference pipelines.}
  \label{fig:analysis:edit}
\end{figure}

\begin{figure}[!t]
  \includegraphics[width=\columnwidth]{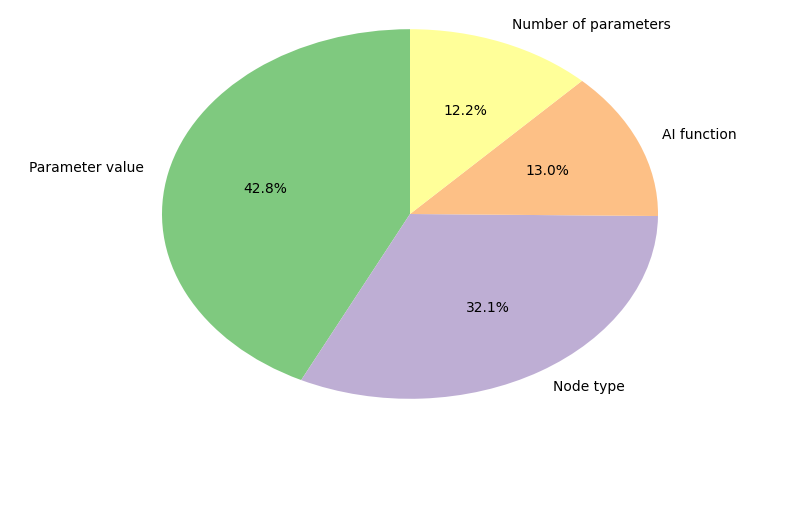}
  \vspace*{-3.5em}
  \caption{Causes for node substitution errors.}
  \label{fig:analysis:subst}
  \vspace{-0.5em}
\end{figure}

\noindent\textbf{Error Types}\hspace{1em}
We analyzed errors in generated pipelines using detailed logs of GED.
Figure \ref{fig:analysis:edit} shows that most errors stem from node substitutions, often due to parameter mismatches or incorrect node types (Figure \ref{fig:analysis:subst}).

Node insertions occur when the builder fails to address all query requirements, often in large pipelines.
Node deletions typically result from redundant function repetitions in separate paths.
Both edits are also required when a misplaced node must be relocated to another path in the graph, which in turn needs corresponding edge insertions and deletions.
These errors are generally less significant compared to node substitutions.

Edge errors often involve missing connections when a function require multiple inputs.
While the Inspector can readily detect these, resolving them remains challenging as it requires comprehensive semantic understanding of the graph and query to locate the correct node supplying the missing data.


\section{Conclusion}
\label{sec:conc}

This paper introduces a novel task of generating AI solution pipelines from user queries and proposes Bel Esprit, a multi-agent framework consisting of Mentalist, Builder, and Inspector, which incrementally improve pipeline quality through query clarification, stepwise construction, and validation.

Future work includes employing retrieval-augmented generation (RAG) with a pool of valid pipelines and extending the framework to generate autonomous agents beyond static pipelines.

\section*{Limitations}
Although the Mentalist (Section \ref{sec:frame:mentalist}) enhances performance in ambiguous scenarios, the system still struggles with highly ambiguous queries, especially when critical input or output requirements are missing.

Pipeline building (Section \ref{sec:frame:builder}) and matchmaking (Section \ref{sec:frame:matchmaker}) are restricted to a predefined pool of AI functions (Appendix \ref{sec:app:function}).
Expanding this pool and incorporating their parameter details increases the prompt length, leading to higher computational costs.
Generic nodes (Section \ref{sec:frame:matchmaker:node}) address this partially but are currently limited to text-to-text functions.

The Inspector (Section \ref{sec:frame:inspector}) does not verify the generated code for script nodes (Section \ref{sec:frame:matchmaker:script}), requiring custom test cases tailored to each script, which is not yet automated.

\bibliography{custom}

\begin{table*}[!t]
\setlength{\aboverulesep}{0pt}
\setlength{\belowrulesep}{0pt}
    \centering
    \renewcommand{\arraystretch}{1.1}
    \scalebox{0.9}{\begin{tabular}{c c c}
        \toprule
        \textbf{Text} & \textbf{Image} & \textbf{Audio} \\
        \cmidrule(lr){1-1} \cmidrule(lr){2-2} \cmidrule(lr){3-3}
        Translation & Image Captioning & Speech Recognition \\
        Summarization & Optical Character Recognition & Speech Synthesis \\
        Text Generation & Document Extraction & Voice Cloning \\
        Text Transformation & Image Generation from Text & Audio Forced Alignment \\
        Question Answering & Image-to-Image Translation & Audio Generation \\
        Text Classification & Image Manipulation & Audio-to-Audio Translation \\
        Topic Classification & Image Classification & Subtitling \\
        Sentiment Analysis & Image Expression Detection & Multilingual Subtitling \\
        Emotion Detection & Object Detection & ASR Quality Estimation \\
        Language Identification & Image Content Moderation & Audio Transcript Analysis \\
        Text Spam Detection & Visual Question Answering & Audio Transcript Improvement \\
        Offensive Language Identification & Depth Estimation & Audio Classification \\
        Text Content Moderation & Image Segmentation & Audio Language Identification \\
        Token Classification & Mask Generation & Audio Speaker Diarization \\
        Named Entity Recognition & Image Compression & Voice Activity Detection \\
        Entity Linking & Image Embedding & Speech Classification \\
        Entity Sentiment Analysis & \textbf{Video} & Speech Embedding \\
        \cmidrule(lr){2-2}
        Coreference Resolution & Video Generation from Text & \textbf{Tabular} \\
        \cmidrule(lr){3-3}
        Syntactic Parsing & Video Generation from Image & Tabular Classification \\
        Semantic Parsing & Viseme Generation & Tabular Captioning \\
        Slot Filling & Extract Audio From Video & Tabular Regression \\
        Text Normalization & Video Speaker Diarization & Table Question Answering \\
        Text Denormalization & Video Classification & Time Series Forecasting \\
        Diacritization & Video Label Detection & \textbf{Others} \\
        \cmidrule(lr){3-3}
        Text Embedding & Video Content Moderation & Similarity Search \\
        & Video Expression Detection & Model Likelihood \\
        \bottomrule
    \end{tabular}}
    \caption{AI functions used in Bel Esprit, categorized by their primary modality.}
    \label{tab:function_list}
\end{table*}

\appendix

\section{List of AI Functions}
\label{sec:app:function}
AI functions in Table \ref{tab:function_list} are considered as possible nodes of a pipeline in this work.

\section{Graph Constraints}
\label{sec:app:graph}

\noindent\textbf{Nodes}\vspace{-0.5em}
\begin{itemize}\itemsep0em
    \item An input node should have no previous nodes\vspace{-0.4em}
    \item An input node should have only one output parameter\vspace{-0.4em}
    \item An output node should have no next nodes\vspace{-0.4em}
    \item There should be no multiple output nodes with the same incoming link\vspace{-0.4em}
    \item A router node should have a single input node as its predecessor\vspace{-0.4em}
    \item A router node should have two or more output parameters, each of which has a different modality\vspace{-0.4em}
    \item A router node should not be connected with another router node\vspace{-0.4em}
    \item A function name should exist in the predefined list of functions\vspace{-0.4em}
    \item Parameters of a function node should exist in the predefined list of parameters\vspace{-0.4em}
    \item A function node should have all its required input parameters
\end{itemize}

\noindent\textbf{Edges}\vspace{-0.5em}
\begin{itemize}\itemsep0em
    \item An input parameter should have only one incoming edge\vspace{-0.4em}
    \item An output parameter should have at least one outgoing edge if it is not an output node\vspace{-0.4em}
    \item Every node should be reachable from an input node\vspace{-0.4em}
    \item An edge should connect existing parameters\vspace{-0.4em}
    \item The connected parameters should have the same modality
\end{itemize}

\begin{figure*}[!ht]
    \centering
    \small
    \begin{tikzpicture}
        \begin{axis}[
            xbar,
            xlabel={Percentage (\%)},
            xmin=0,
            xmax=41,
            ytick=data,
            width=10cm,
            height=4.9cm,
            yticklabels={Educational \& Research, Safety \& Compliance, Information \& Knowledge Management, Content Creation \& Accessibility, Business \& Customer Intelligence, Other},
            nodes near coords,
            nodes near coords align={horizontal},
            nodes near coords style={/pgf/number format/.cd, fixed, fixed zerofill, precision=1}, 
            bar width=10pt,
            enlarge y limits=0.2,
            symbolic y coords={Other, Safety \& Compliance, Education \& Research, Business \& Customer Intelligence, Information \& Knowledge Management, Content Creation \& Accessibility},
        ]
        \addplot[fill=blue!70] coordinates {
            (36.8,Content Creation \& Accessibility)
            (21.4,Information \& Knowledge Management)
            (17.0,Business \& Customer Intelligence)
            (8.8,Education \& Research)
            (8.2,Safety \& Compliance)
            (7.8,Other)  
        };
        \end{axis}
    \end{tikzpicture}
    \caption{Distribution of applications domains of the data entries.}
    \label{fig:data:domain}
\end{figure*}

\section{Query-Pipeline Dataset}
\label{sec:app:data}

\noindent\textbf{Domain Coverage}\hspace{1em} The dataset demonstrates strong coverage of practical applications across various domains (Figure \ref{fig:data:domain}):
\begin{itemize}\itemsep0em
    \item \textbf{Business \& Customer Intelligence}: Analyze company documents or customer feedbacks to gain business insights.
    \begin{itemize}
        \item[$\circ$] \textit{I'm looking for a solution that can identify and categorize customer feedback into different themes, such as product quality, customer service, and delivery experience.}
    \end{itemize}

    \item \textbf{Content Creation \& Accessibility}: Enhance content accessibility across languages and modalities.
    \begin{itemize}
        \item[$\circ$] \textit{I am looking for a solution to convert my French book into an audiobook in the original language as well as in English, Spanish, and Portuguese.}
    \end{itemize}

    \item \textbf{Information \& Knowledge Management}: Extract structured information from unstructured data.
    \begin{itemize}
        \item[$\circ$] \textit{How to generate a 10K rows high-quality Modern Standard Arabic (MSA) corpus for sentiment analysis from an unlabelled text format English dataset?}
    \end{itemize}

    \item \textbf{Safety \& Compliance}: Conduct content moderation and safety applications.
    \begin{itemize}
        \item[$\circ$] \textit{I need a pipeline that can detect and redact sensitive information like personal identifiers from texts, audios, and videos.}
    \end{itemize}

    \item \textbf{Educational \& Research}: Assist students or generate educational materials.
    \begin{itemize}
        \item[$\circ$] \textit{I need a pipeline to assess the readability of documents. The documents are in various languages. Please also provide suggestions for simplification.}
    \end{itemize}
\end{itemize}

\noindent\textbf{Modality Coverage}\hspace{1em} We categorize the task modality of each dataset entry in Figure \ref{fig:data:modality}.
The ``Image \& Video'' and ``Speech \& Audio'' categories include basic transformations to and from text, such as speech recognition.
The ``Multimodal'' category represents more advanced integrations involving multiple modalities, such as functions that process both image and text inputs.

Figure \ref{fig:data:heatmap} illustrates the frequency of modality conversions required to solve the queries in the dataset, showing that all types of transformations between the four modalities are well covered.

\begin{figure}[!t]
    \centering
    \begin{tikzpicture}
        \pie[text=legend, radius=2.3, color={blue!70, red!70, green!70, yellow!70, orange!70}, font=\scriptsize]{35/Text, 28/Image \& Video, 20/Speech \& Audio, 17/Multimodal, 10/Others}
    \end{tikzpicture}
    \caption{Distribution of modalities involved across the data entries.}
    \label{fig:data:modality}
    \vspace{0.5em}
\end{figure}

\begin{figure}[!t]
    \centering
    \small
    \renewcommand{\arraystretch}{1.5}
    \begin{tabular}{| c | >{\centering\arraybackslash}p{3em} >{\centering\arraybackslash}m{3em} >{\centering\arraybackslash}m{3em} >{\centering\arraybackslash}m{3em} |}
        \hline
        \diagbox[width=6em]{Input}{Output} & Text & Audio & Image & Video \\[-0.1em]
    \end{tabular}
    \renewcommand{\arraystretch}{2}
    \begin{tabular}{| >{\centering\arraybackslash}m{4.65em} | >{\centering\arraybackslash}p{3em} >{\centering\arraybackslash}p{3em} >{\centering\arraybackslash}p{3em} >{\centering\arraybackslash}p{3em} |}
        \hline
        Text & \cellcolor{red!80} \scriptsize 25\% & \cellcolor{red!72} \scriptsize 18\% & \cellcolor{red!48} \scriptsize 12\% & \cellcolor{red!32} \scriptsize 8\% \\
        Audio & \cellcolor{red!60} \scriptsize 15\% & \cellcolor{red!40} \scriptsize 10\% & \cellcolor{red!20} \scriptsize 5\% & \cellcolor{red!12} \scriptsize 3\% \\
        Image & \cellcolor{red!56} \scriptsize 14\% & \cellcolor{red!28} \scriptsize 7\% & \cellcolor{red!48} \scriptsize 12\% & \cellcolor{red!16} \scriptsize 4\% \\
        Video & \cellcolor{red!40} \scriptsize 10\% & \cellcolor{red!24} \scriptsize 6\% & \cellcolor{red!20} \scriptsize 5\% & \cellcolor{red!28} \scriptsize 7\% \\
        \hline
    \end{tabular}
    \caption{Heatmap of modality conversions in the dataset's reference pipelines.}
    \label{fig:data:heatmap}
\end{figure}
\end{document}